\documentclass[submission,copyright,creativecommons]{eptcs}
\providecommand{\event}{ICLP 2025} 

\usepackage{iftex}

  \usepackage{underscore}         
  \usepackage[T1]{fontenc}        

\usepackage{mathptmx}
\usepackage{amsmath}
\usepackage{multirow}
\usepackage{amsbsy}
\usepackage{graphicx}
\usepackage{xcolor}
\usepackage{url}
\usepackage{subfigure}
\usepackage{amsfonts}
\usepackage{cprotect}
\usepackage{algorithm}
\usepackage[noend]{algpseudocode}
\usepackage{xspace}

\usepackage{amssymb}
\usepackage{listings}
\usepackage{tikz}
\usepackage{booktabs}
\newcommand{\system}[1]{\ensuremath{\mathtt{#1}}\xspace}

\usepackage{picinpar}
\newtheorem{example}{Example}
\usepackage{makecell}

\usepackage{caption}
\usepackage{wrapfig}
\usepackage{array}

\def\xdnnasp{\system{xDNN^{(ASP)}}}

\def\dnn{\system{DNN}}
\def\deepred{{$\system{DeepRED}$}}

\def\c45{{$\system{C4.5}$}}
\def\newC50{{$\system{C5.0}$}}
\def\eclaire{{$\system{ECLAIRE}$}}
\def\remD{{$\system{REM\text{-}D}$}}
\def\pedc50{{$\system{PedC5.0}$}}

\def\C50{\system{C5.0}}
\def\shap{{$\system{SHAP}$}}
\def\weka{{$\system{WEKA}$}}

\def\naf{{\:\:not\:\:}}
\newcommand{\uffa}{\mbox{$\: \leftarrow \:$}} 
\def\naf{\ensuremath{\mathit{not}}\ \xspace} 

\makeatletter
\lst@Key{countblanklines}{true}[t]{\lstKV@SetIf{#1}\lst@ifcountblanklines}
\lst@AddToHook{OnEmptyLine}{%
    \lst@ifnumberblanklines\else%
    \lst@ifcountblanklines\else%
    \advance\c@lstnumber-\@ne\relax%
    \fi%
    \fi}
\makeatother
\lstdefinelanguage{asp}{
    breakatwhitespace=true,
    captionpos=b,
    numbers=left,
    numbersep=5pt,
    numberblanklines=false,
    countblanklines=false,
    commentstyle=\colour{gray},
    frame=bt, framexbottommargin=5pt, framextopmargin=5pt,
    aboveskip=5pt, belowskip=5pt,
    abovecaptionskip=10pt
}
\lstnewenvironment{asp}{
    \lstset{
        language=asp,
        showstringspaces=false,
        keepspaces=true,
        formfeed=\newpage,
        tabsize=4,
        numbers=none,
        breaklines=true,
        literate={~} {$\sim$}{1},
        frame=none,
    }
}{}

\title{\xdnnasp{}: Explanation Generation System for Deep Neural Networks powered by Answer Set Programming 
}

\author{
Ly Ly Trieu  \qquad\qquad Tran Cao Son
\institute{New Mexico State University, \\
New Mexico, USA}
\email{\quad lytrieu@nmsu.edu \quad\qquad tson@cs.nmsu.edu}
}
\def\titlerunning{\xdnnasp{}: Explanation Generation System for DNN powered by ASP}
\def\authorrunning{L.L. Trieu \& T.C. Son }

\begin{document}
\maketitle

\begin{abstract}
Explainable artificial intelligence (xAI) has gained significant attention in recent years.  
Among other things, explainablility for deep neural networks has been a topic of intensive research 
due to the meteoric rise in prominence of deep neural networks and their ``black-box'' nature.
xAI approaches can be characterized along different dimensions such as their scope (global versus local explanations) or underlying methodologies (statistic-based versus rule-based strategies). Methods generating global explanations aim to provide reasoning process applicable to all possible output classes 
while local explanation methods focus only on a single, specific class.
SHAP (SHapley Additive exPlanations), a well-known statistical technique, identifies important features of a network.
Deep neural network rule extraction method constructs IF-THEN rules that link input conditions to a class. 
Another approach focuses on generating counterfactuals which help explain how small changes to an input can affect the model's predictions.  
However, these techniques primarily focus on the input-output relationship and thus neglect the structure of the network in explanation generation.  

In this work, we propose \xdnnasp{}, an explanation generation system for deep neural networks that provides global explanations. 
Given a neural network model and its training data, \xdnnasp{} extracts a logic program  under answer set semantics that---in the ideal case---represents the trained model, i.e., answer sets of the extracted program correspond one-to-one to input-output pairs of the network.  
We demonstrate experimentally, using two synthetic datasets, that not only the extracted logic program maintains a high-level of accuracy in the prediction task, but it also provides valuable information for the understanding of the model such as the importance of features as well as the impact of hidden nodes on the prediction. The latter can be used as a guide for reducing the number of nodes used in hidden layers, i.e., providing a means for optimizing the network.  
\end{abstract}

\section{Introduction}
Deep neural networks (DNN) have demonstrated their capability to successfully address complex problems 
(see, e.g., \cite{angelov2021explainable,samek2019towards}).
However, the lack of transparency or the absence of an adequate explanation of the results can lead to users' mistrust or skepticism. A major problem, referred as \emph{the problem model/algorithm explainability}, 
is concerned with users' struggle to understand why they receive specific recommendations and/or classification results. 
It is particularly concerning in critical domains such as healthcare 
and cyber-security 
(see, e.g., \cite{angelov2021explainable,ali2023explainable}). 
The significance of explainability has increased with the introduce of  ``the right to an explanation” law by The European Parliament 
(see, e.g., \cite{fandinno2019answering,guidotti2018survey}).
Thus, explainability has become a crucial and essential aspect in the proliferation of deep learning models.  

The study in \cite{guidotti2018survey} classifies xAI methods into \emph{local} and \emph{global} explanations.  
Local explanation refers to the ability of explaining a specific prediction, while global explanation pertains to the overall model, encompassing all predictions. 
A well-know method for local explanations is Local Interpretable Model-Agnostic Explanations (LIME) (\cite{biecek2021local}).
On the other hand, SHapley Additive exPlanation, \shap{} (\cite{lundberg2017unified}), is a model-agnostic, statistically-based method that uses Shapley values to explain feature importance globally. 

Rule extraction methods have also been developed for understanding DNNs. 
Some methods focus on generating counterfactuals which help identifying features that need to be changed to produce a different output (\cite{guidotti2024counterfactual}). These methods allow users to understand how small changes in input can impact the model's decision. 
Some other approaches construct IF-THEN rules from the trained DNN model which are then used to explain the dependence between input feature values and outputs. This provides users with the understanding of the model behavior across all inputs. 
The work in
\cite{ferreira2022looking} is a well-known method that proposes a procedure for inducing logic-based theory for a given neural model (referred to as the main network). The method has to utilize multiple mapping networks that are trained to map the activation values of the main network's output to human-defined concepts. 
The primary cost of this approach lies in the need to label data for the mapping network, which usually requires domain expert knowledge.
Following a taxonomy proposed by 
\cite{ali2023explainable} and \cite{andrews1995survey}
,  rule extraction approaches can be divided into three categories:   
decompositional,  pedagogical and eclectic.
Decompositional methods derive rules based on the structure of the trained neural network and take into consideration hidden layers during extraction. 
On the other hand, pedagogical methods disregard the hierarchical structure in the extraction process. Eclectic methods integrate both decompositional and pedagogical approaches.

It is worth noticing that  some feature/rules extraction methods directly from data 
(\cite{tan2016introduction,pedregosa2011scikit}
such as linear discriminant analysis (LDA) and decision tree could also be used to identify the importance of features and interpretable results. However, these methods do not explain the behavior of the given \dnn{} and have difficulty dealing with irrelevant features. 

In this paper, given a \dnn{} model and an output, we are interested in the following questions: 
(1) ``why is the output produced?", 
(2) ``what features were involved?", and 
(3) ``how was the interaction among the hidden layers?".

To answer the above questions, we propose an \xdnnasp system that belongs to the decompositional xAI group. Different from other existing methods such as \deepred~(\cite{zilke2016deepred}) and \eclaire{}~ (\cite{zarlenga2021efficient}), we take advantage of the answer set programming in producing the set of logic rules in order to provide explanation for answering the three questions. To the best of our knowledge, this is the first work to extract rules in answer set programming from a \dnn{} model. 

\section{Background}
\label{sec:background}

In this work, we will translate \dnn{}s into logic programs  
under the answer set semantics \cite{GelfondL90}.   
A logic program $P$ is a set of rules of the form  \quad 
``$c  \leftarrow a_1,\ldots,a_m,\naf b_{1},\ldots,\naf b_n $'' \quad 
where $c$, $a_i$'s, and $b_j$'s are atoms of a first order language. 
We refer the readers to \cite{GelfondL90} for the precise definition of answer sets of logic programs\footnote{Answer sets can be computed using answer set solvers, 
such as \textit{clingo} \url{https://potassco.org}}. 

Additionally, we utilize the decision tree classification algorithm in Waikato Environment for Knowledge Analysis (\weka{}) as a baseline. \weka{} is a well-known open-source software platform widely used for data mining and machine learning, developed at the University of Waikato in New Zealand \cite{hall2009weka}. It offers simple and user-friendly interfaces that make it easy to create fundamental machine learning models \cite{9359733}.
\weka{} has been referenced widely (see, e.g. \cite{bell2020machine,bhatia2019data,roiger2017data}).
It has also been successfully applied in various domains, including health care and cybersecurity \cite{sharma2019detection,chiu2025waikato}.

\subsection{Deep Neural Network}

A neural network consists of an input layer $i$, an output layer $o$, and multiple hidden layers $h_1,\dots,h_j$, $j \geq 1 $, between them   
(\cite{10.5555/3086952}). 
Each layer consists of a set of nodes, each node is associated with  a bias and an activation function. 
Node $j$ at layer $l-1$ provides an input with weight $w^l_{ij}$ to node 
$i$ at layer $l$.  
The matrices representing the weights and biases of the network are denoted by 
$\textbf{w}$ and $\textbf{b}$, respectively.
The input layer $i$ receives raw features from data, while the output layer $o$ produces the neural network's 
prediction.

The number of nodes in each layer depends on the application and the designer of the network. 
Usually, the number of nodes in the input and output layer corresponds to the number of features in the input data and  
the number of classes of the classification task, respectively.

Given a training set\footnote{
   For simplicity of the representation, we assume a network with one node at the output layer. Multiple output nodes are formalized similarly. 
} $\{(\mathbf{x_i}, y_i) \mid i = 1,...,N\}$, the goal of training a network 
is to learn $\textbf{w}$ to minimize the error function  $ E(\mathbf{w},\mathbf{b}) = \frac{1}{2}\sum_{i=1}^{N}(y_i - \hat{y}_i)^{2}$
where $\hat{y}_i$ represents the output value of the network given the input $\mathbf{x_i}$. 
$a_i^l$ is the activation value at node $i$ at layer $l$, whose activation function is $f$, is computed by 
$a_i^l = f(\sum_j w_{ij}^la_j^{l-1}+b_i^l)$ 
(See, e.g., (\cite{bishop2006pattern}) for more details). 
Different activation functions can be used (e.g., 
\textit{Tanh}, \textit{Sigmoid}, or \textit{ReLU}). 

\subsection{The Algorithm C5.0}
\label{sub:c50}

The algorithm \newC50 was introduced by \cite{quinlan2004data} for a generating classification model from a set of tuples $\{(\mathbf{x_i}, c_i) \mid i = 1,...,N\}$ where $\mathbf{x_i}$ is the attribute values and $c_i$ is its class. 
Compared to its predecessors, \c45 \cite{quinlan1993c4} and ID3 \cite{quinlan1986induction}), \newC50 offers faster processing, better handling of large datasets, and more efficient memory usage.
Detailed formulae for computing these values can be found in the books by \cite{tan2016introduction} and \cite{kuhn2013applied}. For the dataset in Table~\ref{tab:c50}, a set of eight tuples with two attributes 
\emph{h\_2\_n\_0} and \emph{h\_2\_n\_1} and the \emph{xor} class, Algorithm  
\newC50 selects \emph{h\_2\_n\_0} with the threshold \textbf{-0.37244028} and yields a decision tree 
that can be converted into the following two rules with $\mathbf{v}=-0.37244028$: 
\[
\begin{array}{lll}
    (r_1)~  IF ~ h\_2\_n\_0 \leq \mathbf{v} ~ THEN~   xor = 0  &\text{and} &
    (r_2)~ IF ~ h\_2\_n\_0 > \mathbf{v} ~ THEN ~ xor = 1
\end{array}
\]

\begin{wraptable}{l}{0.38\textwidth}
 \centering
 \caption{C5.0 example}
 \label{tab:c50}
 \begin{tabular}{rrc} 
   \\ \hline
\multicolumn{1}{c}{h\_2\_n\_0} & \multicolumn{1}{c}{h\_2\_n\_1} & xor
   \\ \hline
-0.4413446 & 0.6062831 & 0 \\ 
-0.23081768 & -0.6257931 & 1 \\ 
0.5207858 & 0.9767507 & 1 \\ 
\textbf{-0.37244028} & 0.6900585 & 0 \\ 
-0.4413446 & 0.6062831 & 0 \\ 
-0.23081768 & -0.6257931 & 1 \\ 
0.5207858 & 0.9767507 & 1 \\ 
-0.37244028 & 0.6900585 & 0 
   \\ \hline
    \end{tabular}
\end{wraptable} 

Given a set of rules returned by \newC50, a rule $r$'s coverage and correctness, denoted by $r[``cover"]$ and  $r[``ok"]$, respectively, are defined by the number of training instances that satisfy $r$'s condition and that are correctly classified, respectively. 

The confidence value of rule $r$ is defined by 
$\frac{r[``ok"]}{r[``cover"]}*100$. 
The THEN part contains the predicted class, denoted as $r[``class"]$. 
For example, the confidence value of $r_1$ wrt. the data in Table~\ref{tab:c50} is 1  
and $r_1[``class"]=0$.
In the following, we often write 
$t = h\_n\_n\_i \odot v$ to denote the condition between \textit{IF} and \textit{THEN} of a rule $r$, 
where $\odot \in \{\leq,>\}$ and 
use $t[``level"]$, $t[``node"]$, $t[``operation"]$ and $t[``threshold"]$ to refer to $n$, $i$, 
$\odot$, and $v$, respectively.  
For example, for $t = h\_2\_n\_0 \leq -0.37244028$, 
$t[``level"]$, $t[``node"]$, $t[``operation"]$ and $t[``threshold"]$ are 
$2$, $0$, $\leq$, and $-0.37244028$, respectively.

\section{\texorpdfstring{\xdnnasp{}}~ System}
\label{sec:xdnnasp}

\begin{figure}[!th]
    \centering
    \includegraphics[width=.55\linewidth]{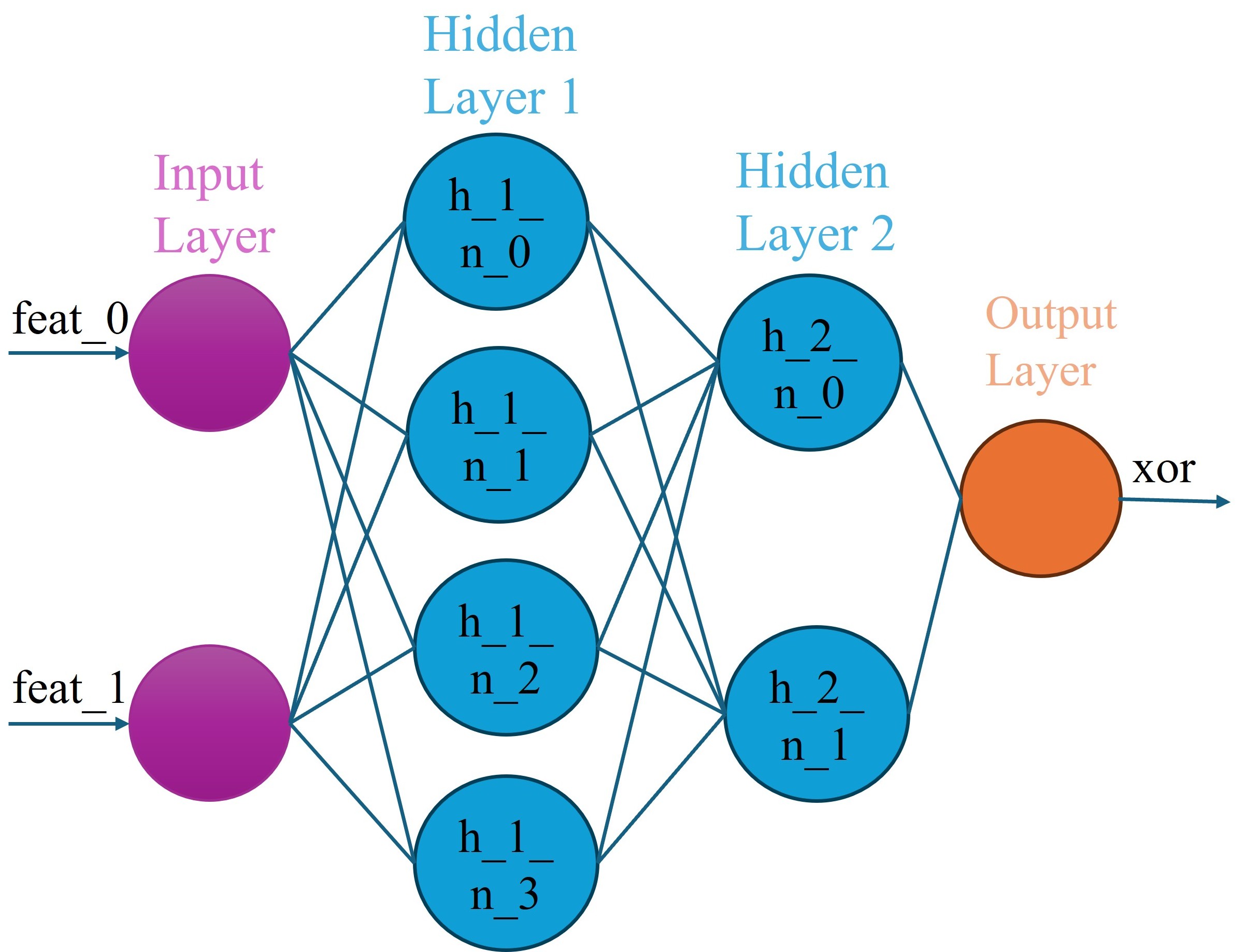}
    \caption{$M_{xor}$: An Example of a \dnn{}.}
    \label{fig:dnnstructure}
\end{figure}

This section describes the $xDNN^{(ASP)}$ system that generates a logic program representation $\Pi(M)$
of a \dnn{} model $M$ whose training data set is $\{(\mathbf{x_i}, y_i) \mid i \in \{1, 2, ...,N\}$.  $\Pi(M)$ will be constructed using the network structure and the activation 
values at each nodes of $M$. 
Assume that $M$ has $k$ hidden layers. 
We write $h_i^j(\textbf{x})$, $1 \le i \le k$, to denote the activation value at node $j$ of layer $i$ for the instance $\textbf{x}$. 
We will begin with a discussion on the atoms and rules of $\Pi(M)$ and continue with the description of the algorithms. For illustration purpose, 
we train a \dnn{} model, $M_{xor}$ (Fig. \ref{fig:dnnstructure}), with 
two binary input features ($feat\_0$ and $feat\_1$), one output XOR,   
two hidden layers (4 nodes in the first hidden layer, 2 in the second), and the \emph{tanh} activation 
function for all nodes. The training set consists of eight tuples, representing the table of the XOR gates, repeated two times\footnote{This is a simplified version of the XOR dataset from \url{https://www.kaggle.com/}, an online platform for data scientists and machine learning professionals,  that repeats the XOR table 2500 times.}
. It is done using 500 epochs, and a batch size of 4. 

\smallskip
\subsection{Atom and Rule Formats}
\label{subsec:atomruleformat}

Given a model $M$, atoms in $\Pi(M)$ correspond to the nodes in $M$. They are classified into three categories: 
\begin{list}{$\bullet$}{\itemsep=0pt \parsep=1pt \topsep=0pt \leftmargin=10pt}
    \item \emph{Input atom} is of the form $input(x, y)$ where $x$ and $y$ 
    is the name and value of a feature; 
    
    \item \emph{Intermediate atom} represents condition associated with a node $j$ in hidden layer $i$ and takes one of the forms:  
    \quad $h(i,j,``\leq",v,true) \quad \textnormal{or} \quad  
        h(i,j,``>",v,true)
    $  \\ 
    where $v$ is the threshold associated to the rule that is generated by \newC50 and contains 
    the condition related to node $j$ at layer $i$, 
    e.g., $ h\_2\_n\_0 \leq -0.37244028$ in rule $r_1$ in Section~\ref{sub:c50} is 
    represented as  
    $h(2,0,``\leq",``-0.37244028",true).$
      
    \item \emph{Class atom} is encoded using predicate $potential\_predict\_output(x, y, z)$ where $x$ denotes the predicted class label, $y$ is the identifier of the rule, and $z$ is a confidence value associated with the prediction in the rule.
\end{list}
Rules in $\Pi(M)$ encode the relationships between nodes. For a rule $head \leftarrow body$ in $\Pi(M)$, 
if the $head$ is an atom corresponds to a node at level $i$ of $M$, where $i=0$ ($i=k+1$) represents the input (output)  layer, then the $body$ can only contain literals corresponding to nodes at level $i-1$. 
For ease of the presentation, rules are organized into three levels corresponding to the three types of connections in the neural network (input-to-hidden, hidden-to-hidden, and hidden-to-output), as follows: (\emph{i}) \emph{Top-level} rules have a class atom as the head and the intermediate literals from the last hidden layer as the body; 
(\emph{ii}) \emph{Intermediate rules} are those whose head and body contain only intermediate literals.
(\emph{iii}) \emph{Bottom-level rules} have an intermediate literal from the first hidden layer as the head. 
    To simplify the experiment, we construct a bottom-level rule with 
    a body consists of (\emph{i}) input atoms and (\emph{ii}) arithmetic comparisons between the variable in the second term of an input atom and a threshold $v$ derived from the trained model.

\subsection{Algorithms of \texorpdfstring{\xdnnasp{}}~}
\label{subsec:xdnnasp}

\begin{algorithm}[!th]
\caption{$xDNN^{(ASP)}$}
\label{alg:xdnnasp}
\textbf{Input:} a \dnn{} model $M$ with hidden layers $\{h_1, ..., h_k\}$ and \\
\hspace*{1cm}\: training dataset $X = \{(\textbf{x}^{(1)},y_1), (\textbf{x}^{(2)},y_2), .., (\textbf{x}^{(N)},y_N)\}$\\
\textbf{Output:} Answer set program $\Pi(M)$
\begin{algorithmic}[1]

\State $data_{top} \gets \{(h_k(\textbf{x}^{(1)}),y_1),(h_k(\textbf{x}^{(2)}),y_2), ..., (h_k(\textbf{x}^{(N)}),y_N)\}$
\State $\Pi(M)_{top},Conditions \gets extract\_top\_level(data_{top})$
\State $\Pi(M) \gets \Pi(M)_{top}$
\For {hidden layer $i = k-1,k-2,..,1$}
    \State $tem\_Conditions \gets \emptyset$
    \For {$t \in Conditions$}
        \State $h_{level}, n_{index}, \odot, th \gets t[``level"], t[``node"], t[``operator"], t[``threshold"]$
        \State $y_1',y_2', ..., y_N' \gets t(h_{h_{level}}^{n_{index}}(\textbf{x}^{(1)})),t(h_{h_{level}}^{n_{index}}(\textbf{x}^{(2)})),..,t(h_{h_{level}}^{n_{index}}(\textbf{x}^{(N)}))$
        \State $data_{i}^{t} \gets \{(h_i(\textbf{x}^{(1)}),y_1'),(h_i(\textbf{x}^{(2)}),y_2'), ..., (h_i(\textbf{x}^{(N)}),y_N')\}$

        \State $\Pi(M)_{i}^t,Conditions_{i}^t \gets extract\_intermediate\_level(data_{i}^{t},h_{level},n_{index},\odot,th)$
        
        \State $\Pi(M) \gets \Pi(M) \cup \Pi(M)_{i}^t$
        \State $tem\_Conditions \gets tem\_Conditions \cup Conditions_{i}^t$
        
    \EndFor
    \State $Conditions \gets tem\_Conditions$
\EndFor
\For {$t \in Conditions$}
    \State $1, n_{index}, \odot, th \gets t[``level"], t[``node"], t[``operator"], t[``threshold"]$

    \State $y_1',y_2', ..., y_N' \gets t(h_{1}^{n_{index}}(\textbf{x}^{(1)})),t(h_{1}^{n_{index}}(\textbf{x}^{(2)})),..,t(h_{1}^{n_{index}}(\textbf{x}^{(N)}))$

    \State $data_{input}^{t} \gets \{(\textbf{x}^{(1)},y_1'),(\textbf{x}^{(2)},y_2'), ..., (\textbf{x}^{(N)},y_N')\}$

    \State $\Pi(M)_{bottom}^t\gets extract\_bottom\_level(data_{input}^{t},1,n_{index},\odot,th)$
    
    \State $\Pi(M) \gets \Pi(M) \cup \Pi(M)_{bottom}^t$
\EndFor
\State \Return $\Pi(M)$
\end{algorithmic}
\end{algorithm}
Algorithm \ref{alg:xdnnasp} is the main algorithm for constructing $\Pi(M)$ given $M$ 
with $k$ hidden layers and its training dataset 
 $X = \{(\textbf{x}^{(1)},y_1), (\textbf{x}^{(2)},y_2), .., (\textbf{x}^{(N)},y_N)\}$. 
 It works as follows. 
\begin{list}{$\bullet$}{\itemsep=0pt \parsep=1pt \topsep=0pt \leftmargin=10pt}
    \item \emph{Computing top-level rules, $\Pi(M)_{top}$:}  
    It starts with computing the activation values for the layer $k$ (Line 1).
    Algorithm~\ref{alg:extracttop} is invoked to compute 
    $\Pi(M)_{top}$ together with the set of conditions in the rules of $\Pi(M)_{top}$ (Line 2). 

    \item \emph{Computing intermediate-level rules} (Lines 4-13).
    This is done through a two-stage iterative process.
    First, it iterates over each hidden layer $i$ (Line 4).
    Then, it processes each condition $t$ in $Conditions$, 
    it identifies the level, node index, operator, and threshold from $t$ (Line 7).
    Furthermore, it evaluates the condition $t$ given the activation value at node 
    given the input values $\textbf{x}^{(1)}$, $\textbf{x}^{(2)}$, \dots,$\textbf{x}^{(N)}$ 
    (Line 8) and prepares the data for the extraction of the 
    intermediate-rules (Line 9). Finally, it calls Algorithm \ref{alg:extractintermediate} 
    to compute the intermediate rules pertaining to $t$ ($\Pi(M)^t_i$) and the set of conditions 
    occurring in this set of rules ($Conditions^t_i$) (Line 10). 
    Lines 11-12 do some housekeeping and it is obvious what they are doing.  
    
    \item  \emph{Computing bottom-level rules, $\Pi(M)_{bottom}$} (Lines 14-19). 
    After completing the final iteration for computing intermediate-level rules with $i=1$, 
    for each condition $t$ that occurs in the first hidden layer,  
    the level with a value of 1, node index, operator, and threshold is extracted (Line 15).
    The evaluation of $t$ given the training data set is computed (Line 16), 
    which is then used to create the data for the bottom-rule extraction process (Line 17).  
    The algorithm calls 
    Algorithm \ref{alg:extractbottom} to compute the set of bottom-rules associated with $t$, $\Pi(M)_{bottom}^t$ (Line 19) which is 
    added to $\Pi(M)$. 

\end{list}

\subsubsection{Computing Top-Level Rules}

\begin{algorithm}[!th]
\caption{$extract\_top\_level(data_{top})$}
\label{alg:extracttop}
\begin{algorithmic}[1] 

\State Initializing: $\Pi(M)_{top} \gets \emptyset$; \quad  $Conditions \gets \emptyset$; \quad   $index \gets 0$
\State $R_{top} \gets C50(data_{top})$
\For {$r \in R_{top}$}

    \State $T \gets getConditions(r)$; \quad  $c \gets r[``ok"]/r[``cover"]$

    \State $head \gets potential\_predict\_output(r[``class"],index,c)$; \quad   $body \gets \emptyset$

    \For {$t \in T$}
        \State $Conditions,literal \gets process\_condition(t,Conditions)$
        \State $body \gets body \cup \{literal\}$
    \EndFor
    \State $\Pi(M)_{top} \gets \Pi(M)_{top} \cup \{head \leftarrow body\}$; \quad $index \gets index + 1$

\EndFor
\State \Return $\Pi(M)_{top},Conditions$
\end{algorithmic}
\end{algorithm}

\paragraph{Algorithm \ref{alg:extracttop}} utilizes \newC50 to compute a set of  IF-THEN rules $R_{top}$ (Line 2) given the data set  $data_{top}$.   
Each rule $r$ is associated with an 
$index$ (initialized in Line 1 and incremented in Line 9) and 
consists of a set of arithmetic conditions  $T$ (Line 4), a predicted class label ($r[``class"]$), 
and a confidence value $c$ (Line 4). 
This is used to constructed the class atom $head$ of a top-level rule (Line 5) . 
The body of this rule is computed in Lines 6--8 by processing each term $t$ in $T$ via Algorithm \ref{alg:processcondition} that returns  the literal needed to be in the body and the set $Conditions$ (Line 7). 
The meaning of Lines 9-10 are obvious.

\smallskip \noindent  
\textbf{Algorithm \ref{alg:processcondition}} receives an arithmetic condition $t$ 
(result from \newC50)
and a set of 
conditions $Condition$ and processes $t$ by identifying the operator in $t$ as follows: (\emph{i})
updates $Condition$ if $t$ and its opposite condition do not belong to $Conditions$ (Lines 7--8); 
and (\emph{ii}) creates a literal $l$ for use in the body of top-level rules (Lines 9-11). 
Note that $l$ can be of the form ``\emph{not h(i,j,$\odot$,value,``true'')}'' (Line 11),
leveraging the existing condition. This reduces the number of newly created atoms, as they can be described with the existing ones.
We illustrate the execution of Algorithm~\ref{alg:extracttop} in the next example.

\begin{algorithm}[!ht]
\caption{$process\_condition(t,Conditions)$}
\label{alg:processcondition}
\begin{algorithmic}[1] 

\State $h_{level}, n_{index}, \odot, th \gets t[``level"], t[``node"], t[``operator"], t[``threshold"]$
\If {$\odot = ``\leq"$}
    \State $\ominus  \gets ``>"$
\Else
    \State $\ominus \gets ``\leq"$
\EndIf
\State $opposite\_t \gets h_{level},n_{index},\ominus,th$
\If {$t \notin Conditions \land opposite\_t \notin Conditions$}
    \State $Conditions \gets Conditions \cup \{t\}$
\EndIf

\State $l \gets concatenate(h_{level}, n_{index}, \odot, th,``true")$

\If{$opposite\_t \in Conditions$}
    \State $l \gets concatenate(``not\hspace{2mm}",h_{level}, n_{index},\ominus,th,``true")$
\EndIf

\State \Return $Conditions,l$
\end{algorithmic}
\end{algorithm}

\begin{example}
\label{ex:exforhiddenouput}
    Given the trained model $M_{xor}$ and its training set as input to Algorithm \ref{alg:xdnnasp},
    Table \ref{tab:c50} 
    is the dataset $data_{top}$ computed in Line 2 of Algorithm \ref{alg:xdnnasp}. 
    \newC50 generates two rules in $R_{top} = \{r_1,r_2\}$ (Line 2, Algorithm \ref{alg:extracttop}). 
    
    For $r = r_1$, $T$ contains only one condition $t = h\_2\_n\_0 \leq -0.37244028$, which is added to the $Conditions$. We obtain the first logic rule as follows:
    \[
    (r'_1):
    potential\_predict\_output(0,0,1) \leftarrow h(2,0,``\leq",``-372440",true). 
    \]
 
    For $r = r_2$, $T$ contains $t = h\_2\_n\_0 > -0.37244028$. Because its opposite condition belongs to $Conditions$, the default negative atom of the existing condition is used, resulting a second rule as follows:
    \[
    (r'_2):
    potential\_predict\_output(1,1,1) \leftarrow \naf h(2,0,``\leq",``-372440",true).
    \]
    Algorithm \ref{alg:extracttop} returns $\Pi(M)_{top} = \{r'_1,r'_2\}$ and $Conditions=\{h\_2\_n\_0 \leq -0.37244028\}$. 
\end{example}

\subsubsection{Computing Intermediate Rules}

\begin{algorithm}[!th]
\caption{$extract\_intermediate\_level(data_{i}^{t},h_{level},n_{index},\odot,th)$}
\label{alg:extractintermediate}
\begin{algorithmic}[1] 

\State Initializing: $\Pi(M)_{i}^t \gets \emptyset$; \quad 
  $Conditions_{i}^t \gets \emptyset$; \quad 
  $R_{i}^{t} \gets C50(data_{i}^{t})$
\For {$r \in R_{i}^{t}$}
    \State $T \gets getConditions(r)$
    \State $class \gets r[``class"]$
    \If {$class = True$}
        \State $head \gets concatenate(h_{level},n_{index},\odot,th,class)$
        \State $body \gets \emptyset$
        
        \For {$t \in T$}
            \State $Conditions_{i}^t,literal \gets process\_condition(t,Conditions_{i}^t)$
            \State $body \gets body \cup \{literal\}$
        \EndFor
        \State $\Pi(M)_{i}^t \gets \Pi(M)_{i}^t \cup \{head \uffa body\}$
    
    \EndIf
\EndFor
\State \Return $\Pi(M)_{i}^t,Conditions_{i}^t$
\end{algorithmic}
\end{algorithm}
\noindent
\textbf{Algorithm \ref{alg:extractintermediate}} computes the set of intermediate rules 
representing the dependencies between nodes in layer $i$ and layer $i+1$. 
Similar to Algorithm~\ref{alg:extracttop}, it invokes \newC50 to generate a set of rules $R_i^t$ from the input set of activation values and constructs the intermediate rules from $R_i^t$. The key difference between this algorithm and Algorithm~\ref{alg:extracttop} lies in that it only processes rules in $R_i^t$ whose prediction is the true class (Lines 5--11). 

\begin{table}[ht!]
 \centering
 \caption{
 $data_{1}^{t}$ for $t = h\_2\_n\_0 \leq -0.37244028$ (Algorithm~\ref{alg:extractintermediate})}
 \label{tab:exintermediatelayer}
 \begin{tabular}{rrrrc} 
   \\ \hline
\multicolumn{1}{c}{$h\_1\_n\_0$} & \multicolumn{1}{c}{$h\_1\_n\_1$} & \multicolumn{1}{c}{$h\_1\_n\_2$} & \multicolumn{1}{c}{$h\_1\_n\_3$} & 
$h\_2\_n\_0\_leq\_than\_minus372440$ 
\\
  \midrule
 -0.33337042 & 0.22323501 & -0.35911334 & -0.30652052 & True\\ 
 -0.93152505 & 0.74552625 & 0.43851086 & -0.88826376 & False\\ 
 0.41850552 & 0.7016869 & -0.83853734 & 0.4565624 & False\\ 
 -0.70506656 & 0.92262095 & -0.35396752 & -0.5398724 & True\\ 
 -0.33337042 & 0.22323501 & -0.35911334 & -0.30652052 & True\\ 
 -0.93152505 & 0.74552625 & 0.43851086 & -0.88826376 & False\\ 
 0.41850552 & 0.7016869 & -0.83853734 & 0.4565624 & False\\ 
 -0.70506656 & 0.92262095 & -0.35396752 & -0.5398724 & True
   \\ \hline
    \end{tabular}

\end{table}

\begin{example}
\label{ex:hiddenhiden}
    Continuing from Example \ref{ex:exforhiddenouput}, Table \ref{tab:exintermediatelayer} presents the dataset $data_1^t$, where $t=h\_2\_n\_0 \leq -0.37244028$. 
    \C50  uses $data_1^t$ to generate the following three IF-THEN rules:
    \vspace{-5pt}
\[
\begin{array}{lclll}
    (r_3)  & IF~ h\_1\_n\_1 > 0.22323501  \land h\_1\_n\_1 \leq 0.74552625  & ~THEN~ class = False, \\ 
    (r_4)  & IF~ h\_1\_n\_1 \leq 0.22323501 & ~THEN~ class = True, \\
    (r_5)  & IF~ h\_1\_n\_1 >  0.74552625 & ~THEN~ class = True
\end{array}
\]   

\smallskip 
\indent Algorithm~\ref{alg:extractintermediate} processes only $r_4$ and $r_5$ and 
and returns $\Pi(M)_{1}^t = \{r'_4,r'_5\}$ and $Conditions_{1}^t = \{h\_1\_n\_1 \\ \leq 0.22323501, h\_1\_n\_1 >  0.74552625\}$ where 
\[
\begin{array}{lllll}
(r'_4)  & h(2,0,``\leq",``-372440",true) \uffa h(1,1,``\leq",223235,true). \\
(r'_5) & 
h(2,0,``\leq",``-372440",true) \uffa  h(1,1,``>",745526,true).
\end{array}
\]   
\end{example}

\subsubsection{Computing Bottom-Rules}

\paragraph{\textbf{Algorithm \ref{alg:extractbottom}}} computes the set of bottom rules $\Pi(M)_{bottom}^t$ 
and is similar to Algorithm~\ref{alg:extractintermediate} in that it invokes \newC50 to compute the set $R$ of IF-THEN rules and processes only rules that predict a True class (Line 5) to corresponding bottom-rules in $\Pi(M)_{bottom}^t$. The process of each condition of each rule in $R$ creates rules with variables representing possible pairs of features and values (Lines 10-11) to enable the evaluation of the program using all possible inputs of a feature. 

\begin{algorithm}[!th]
\caption{$extract\_bottom\_level(data_{input}^{t},1,n_{index},\odot,th)$}
\label{alg:extractbottom}
\begin{algorithmic}[1] 
\State Initializing: $\Pi(M)_{bottom}^t \gets \emptyset$; \quad 
  $R \gets C50(data_{input}^{t})$
\For {$r \in R$}
    \State $T \gets getConditions(r)$
    \State $class \gets r[``class"]$
    \If {$class = True$}
        \State $head \gets concatenate(1,n_{index},\odot,th,class)$
        \State $body \gets \emptyset$
        \For {$t \in T$}
            \State $h_{level}, n_{index}, \odot, th \gets t[``level"], t[``node"], t[``operator"], t[``threshold"]$
            \State $atom \gets input(h_{level} n_{index},``Value"h_{level}n_{index})$
            \State $arithmetic\_comparison \gets concatenate(``Value"h_{level}n_{index},\odot,th)$
            \State $body \gets body \cup \{atom,arithmetic\_comparison\}$
        \EndFor
        \State $\Pi(M)_{bottom}^t \gets \Pi(M)_{bottom}^t \cup \{head \uffa body\}$
    
    \EndIf
\EndFor
\State \Return $\Pi(M)_{bottom}^t$
\end{algorithmic}
\end{algorithm}

\begin{example}
The dataset $data_{input}^t$ for $t=h\_1\_n\_1 \leq 0.22323501$ is encoded as a set of eight pairs $(x_i,y_i)$, where $x_i=(input\_feat\_0,input\_feat\_1)$ and the label $y_i$ is named $h\_1\_n\_1\_leq\_than223235$ as follows: 
$ \{((0,0),True),
((0,1),False),
((1,0),False),
((1,1),False), 
((0,0),True),
((0,1),False),\\
((1,0),False),
((1,1),False),
\}$.    
\newC50 generates the following rules:   

\[
\begin{array}{llclll}
    (r_6) & IF~ &input\_feat\_1 > 0 & ~THEN~ & class = False, \\
    (r_7) & IF~ & input\_feat\_0 > 0 &   ~THEN~ &class = False, \\   
    (r_8) & IF~ &input\_feat\_0 \leq 0 \land input\_feat\_1 \leq 0 & ~THEN~ & class = True 
\end{array}
\]

Only rule $r_8$ is processed by Algorithm~\ref{alg:extractbottom}  and produces the  
following rules:

\[
\begin{array}{lrr}
   (r'_8) \quad h(1,1,``\leq",223235,true) \uffa & & \\
   \qquad input(input\_feat\_0,Value\_input\_feat\_0),  
    Value\_input\_feat\_0 \leq 0, && \\
   \qquad input(input\_feat\_1,Value\_input\_feat\_1), 
     Value\_input\_feat\_1 \leq 0. &&
\end{array}
\] 

Due to space limitation, the process for condition $t'= h\_1\_n\_1 >  0.74552625$ is 
omitted. We note that the following rule is produced by Algorithm~\ref{alg:extractbottom}: 

\[
\begin{array}{lrr}
   (r'_9) \quad h(1,1,``>",745526,true) \uffa & & \\
   \qquad input(input\_feat\_0,Value\_input\_feat\_0), Value\_input\_feat\_0 > 0,  && \\
   \qquad input(input\_feat\_1,Value\_input\_feat\_1),  Value\_input\_feat\_1 > 0. &&
\end{array}
\] 

To summarize,
Algorithm \ref{alg:xdnnasp} returns $\Pi(M_{xor}) = \{r'_1,r'_2,r'_4,r'_5,r'_8,r'_9\}$. Assuming no noise occurs,
observe that  
    \begin{itemize}
        \item Class 0 is predicted if the condition at node 0 in hidden layer 2 is True, with 100$\%$ confident value ($r'_1$). This occurs  when at least one of the conditions at node 1 in hidden layer 1 is True ($r'_4$ and $r'_5$), and this happens when the feature 0 and feature 1 of a given instance are either both $\leq 0$ ($r'_8$) or both $> 0$ ($r'_9$).
        \item Class 1 is predicted if there is no evidence that the condition at node 0 in hidden layer 2 is True, with 100$\%$ confident value ($r'_2$). This happens when both conditions at node 1 in hidden layer 1 are not applicable ($r'_4$ and $r'_5$), meaning that the feature 0 and feature 1 of a given instance are not simultaneously $\leq 0$ or $>0$. 
    \end{itemize}

Before presenting our experiments, let us remark that by construction,  
$\Pi(M)$ is stratified.  
This facilitates the proof that given an input $\mathbf{x} = \mathbf{v}$ in the form of a set of atoms $I = \{input(x_i, v_i) \mid x_i$ is a feature of $\mathbf{x}\}$, the program $\Pi(M) \cup I$ has a unique answer set. Furthermore, the size of the program  grows proportionally to the size of the network.  

\end{example}

\section{Experiments}
\label{sec:experiment}

In this section, we evaluate our system on two synthetic datasets.
The first dataset, named \emph{XOR}, evaluates the performance of the output of \xdnnasp{} in prediction, consistency, and identifying the feature importance. It contains 1000 samples, each with 10 features, 
originated in the work of \cite{zarlenga2021efficient}.
The 10-dimensional feature vectors are generated from a uniform distribution in $[0, 1]$, and the binary class is $y_i = round(\textbf{x}^{(i)}_1) \oplus round(\textbf{x}^{(i)}_2)$, where $\oplus$ presents the XOR operation between the $1^{st}$ and $2^{nd}$ features. 
The second dataset, named \emph{Modified-XOR}, is a slight modification of the \emph{XOR} dataset where the 
class label is defined by $y_i = (round(\textbf{x}^{(i)}_1) \oplus round(\textbf{x}^{(i)}_2)) \oplus round(\textbf{x}^{(i)}_3)$.
We note that for \emph{XOR}, only $x^i_1$ and $x^i_2$ contribute to the value of the output, i.e., only 
two  features  are important. 
On the other hand, for \emph{Modified-XOR}, three features, $x^i_1$, $x^i_2$, and $x^i_3$ are important.
This means that \emph{Modified-XOR} is more complex than \emph{XOR}.
\emph{Modified-XOR} was designed to investigate the influence of hidden nodes, which could contribute to the redesign of neural network architectures.

Experiments explored various neural network configurations such as the number of hidden layer nodes $\{8,16,32,64,128\}$,  potential activation functions  $\{tanh,relu,elu\}$, number of epochs  $\{50,100,\\150,200\}$, and batch size  $\{16,32\}$. 
The input layer had 10 nodes, and the output layer for binary classification had 1 node, following \cite{tan2016introduction}.

Recall that \xdnnasp{} computes $\Pi(M)$ given a trained \dnn{} model $M$ and its training dataset $X$.
Given an instance $(\mathbf{x},y)$, we say that $\Pi(M)$ correctly predicts the class $y$ if $y$ 
is a \emph{most appropriate class} in the answer set $A$ of $\Pi(M) \cup input(x_i,v_i)$ 
where $y$ is a most appropriate class if it is the maximal element of the ordering $>$ 
defined by $l > l'$ if 
(\emph{i}) 
the number of atoms of the form 
$potential\_predict\_output(l, \_, \_)$ is larger than the number of atoms 
of the form 
$potential\_predict\_output(l', \_, \_)$; 
or (\emph{ii}) the number of atoms of the form 
$potential\_predict\_output(l, \_, \_)$ equals the number of atoms 
of the form 
$potential\_predict\_output(l', \_, \_)$ 
and $l$ occurs in an atom with a higher confidence value.  
The accuracy of the program $\Pi(M)$ is defined as the percentage of the number of corrected prediction over the size of the test dataset. It reflects how well is $\Pi(M)$ comparing to $M$  
(see Tables \ref{tab:exxor} and \ref{tab:exmodifiedxor}). 
Fidelity measures the percentage between the accuracy of the extracted rules and the original model from which the rules were derived, showing how closely our extracted rules reflect the performance of the original model 
(see Figures \ref{fig:ourmethodotherxor} and \ref{fig:eclareourmethodxor}).

We compare \xdnnasp{} with \eclaire{}, the best performer among the decompositional algorithms 
\deepred{} (\cite{zilke2016deepred}), \remD{} (\cite{shams2021rem}) and \eclaire{}  (\cite{zarlenga2021efficient}). 
The key distinction between \xdnnasp{} and \eclaire{} lies in that \eclaire{} only generates a set of 
IF-THEN rules that map input feature conditions directly to a class. We do not include a comparison \xdnnasp{}  
\shap{} (\cite{lundberg2017unified}) in this paper 
as the latter does not extract rules describing input-output dependencies. We note that in our study,  
our method and SHAP identify the same top features.

\subsection{Accuracy, Consistency and Feature Importance}
\label{subsec:exxor}

\begin{table}[th]
 \centering
 \caption{$M$ vs. $\Pi(M)$ in 5 configurations}
 \label{tab:exxor}
 \begin{tabular}{llllllr}
   \\ \hline
     Design\: & \makecell{Hidden} \: & Epoch \: & \makecell{Batch} \: &
  \multicolumn{3}{c}{Accuracy ($\%$)}  \\
        \cmidrule{5-7} 
    &nodes & size & & $M$ &\:\: & $\Pi(M)$  
     \\ \hline
    1  & 128 &  200 & 32 & 96.3 & & 94   \\
    2  & 128 &  150 & 32 & 95.2 & & 94   \\
    3  & 64 &  200 & 16 & 96.2 & & 93.8   \\
   4   & 128 &  150 & 16 & 95.5 & & 93.2   \\
   5   & 32 &  200 & 16 & 95.1 & & 92.6  
   \\ \hline
    \end{tabular}

\end{table}

To establish a baseline for our experiment, we use Weka (\cite{hall2009weka}) and create 
decision tree classification using 5-fold cross-validation. This baseline has accuracy of 84.4$\%$.
We then experiment with different configurations for the \dnn{} used in this experiment. The reported designs in Table~\ref{tab:exxor} are the five with accuracy over 90\%. In this experiment, we use a simpler network with one hidden layer.
All NNs achieved over 95$\%$ accuracy with 5-fold stratified cross-validation.
\xdnnasp{} achieved the highest rule extraction accuracy in designs 1 and 2 at 94$\%$, 
while design 5 (with 32 hidden nodes) had the lowest rule accuracy at 92.6$\%$ among them.
Our results are comparable to \eclaire{}, which achieve $91.8 \pm 2.6\%$  rule accuracy using  a neural network with 3 hidden layers (\cite{zarlenga2021efficient}).

\begin{figure}[!ht]
    \centering
    \includegraphics[width=.55\linewidth]{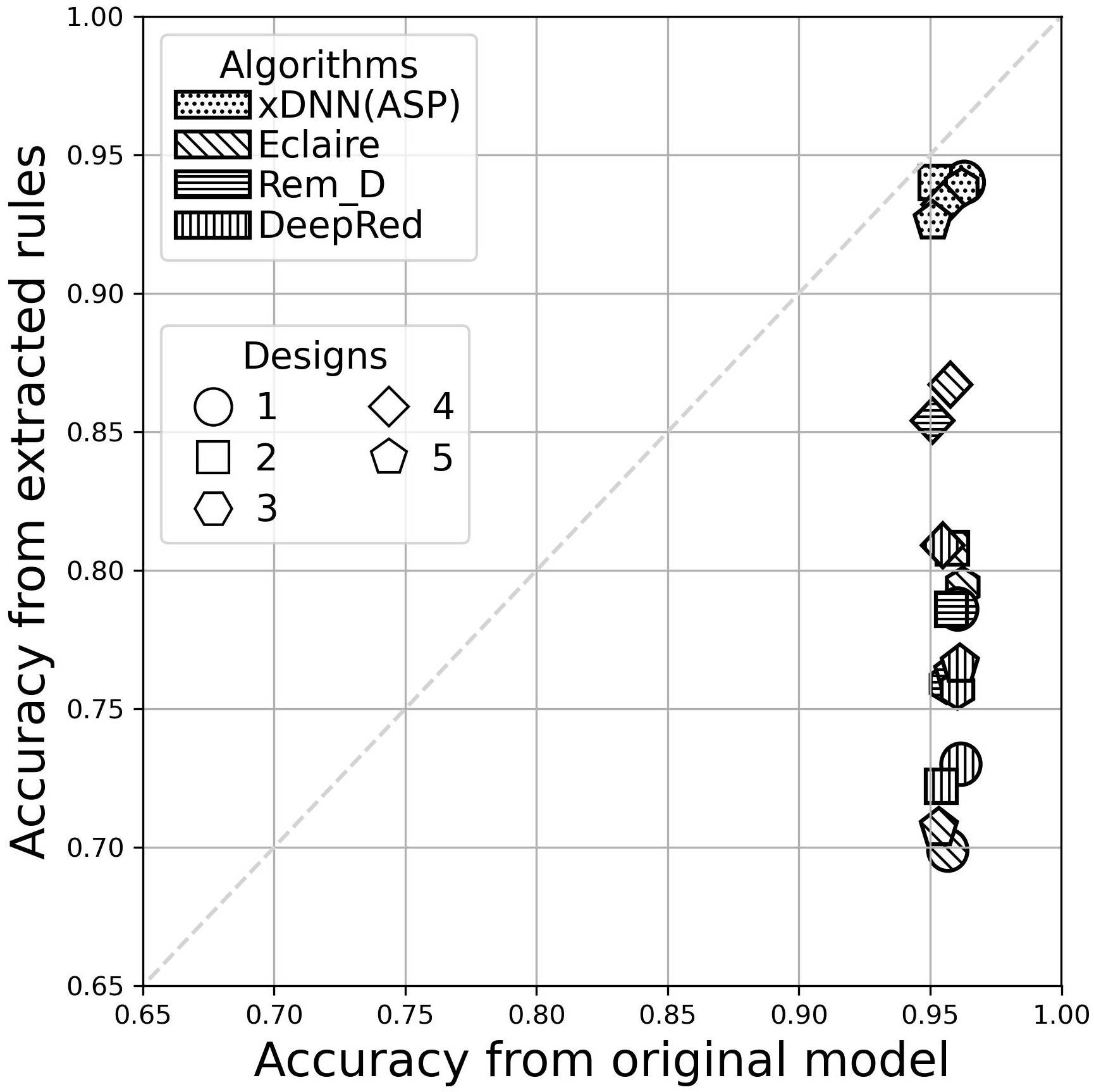}
    \caption{Comparing  \xdnnasp{}, \eclaire{}, \remD{} and \deepred{}.}
    \label{fig:ourmethodotherxor}
\end{figure}

\begin{figure}[!th]
    \centering
    \includegraphics[width=.55\linewidth]{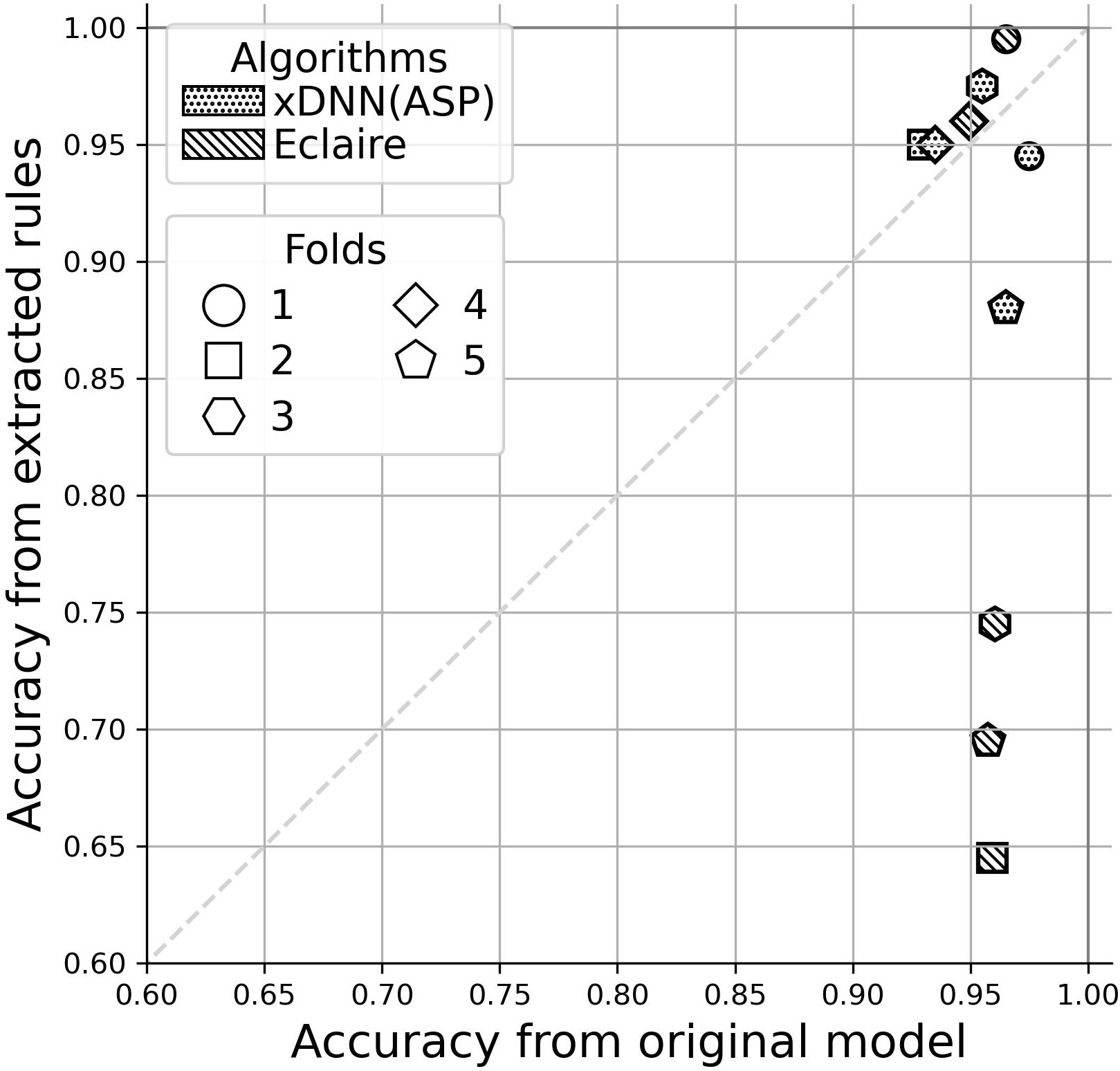}
    \caption{\xdnnasp{} vs. \eclaire{} wrt. 5-fold in Design 2, Table \ref{tab:exxor}.}
    \label{fig:eclareourmethodxor}
\end{figure}

\begin{figure}[!ht]
    \centering
    \includegraphics[width=.95\linewidth]{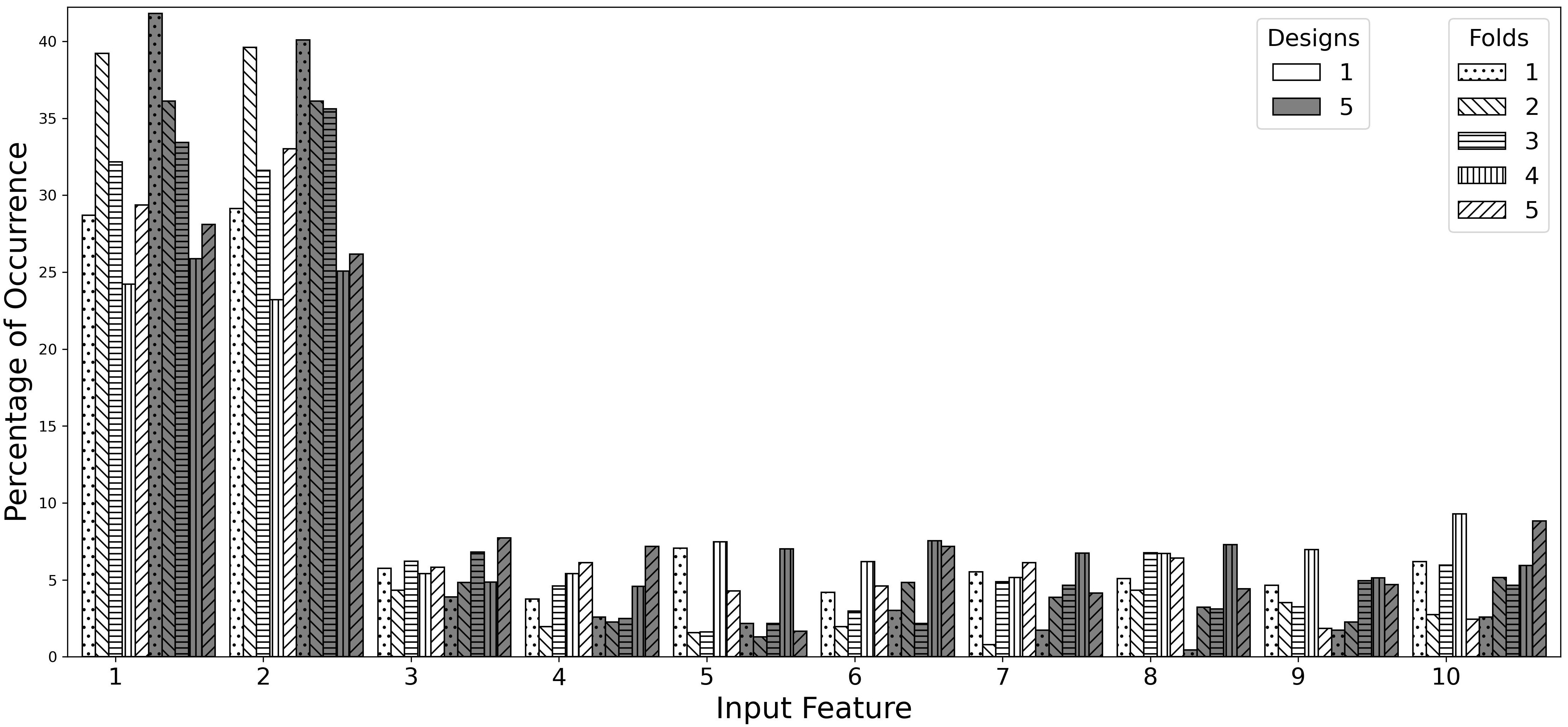}
    \caption{Feature importance in $\Pi(M)$ across 5 folds in the designs 1 and 2  from Table \ref{tab:exxor}}
    \label{fig:featureimportancexor}
\end{figure}

Fig. \ref{fig:ourmethodotherxor} compares the accuracy of NNs and four rule extraction algorithms across five designs in Table \ref{tab:exxor}, while Fig. \ref{fig:eclareourmethodxor} presents fold-level accuracy  for NN, \eclaire{} and \xdnnasp{} on design 2. 
If the extracted rules closely mirror the behavior of their original NN models, their accuracies should align closely, meaning that each point should lie near the diagonal line. 
In Fig. \ref{fig:ourmethodotherxor},  while all NN models performed well across the five designs, 
the rule accuracy of \xdnnasp{} consistently outperformed the others, demonstrating high fidelity. 
Fig. \ref{fig:eclareourmethodxor}  shows that, while \eclaire{} reached 99.5$\%$ accuracy in one fold, its performance dropped to about 65$\%$ in another, indicating inconsistency within the same design. 
Thus, \eclaire{} may need more exploration of NN designs to improve both rule extraction and model accuracy.
In contrast, \xdnnasp{} delivered more stable results across all folds, despite slightly lower peak accuracy.
We define the importance of a feature $x_i$ as the percentage between the number of occurrences of $input(x_i,\_)$ in $\Pi(M)$ compared to the total number of occurrences of all features in $\Pi(M)$.
Fig. \ref{fig:featureimportancexor} shows these percentages for each input feature in the extracted rules across five folds for design 1 (highest accuracy) and design 5 (lowest accuracy).  The $1^{st}$ and $2^{nd}$ features contribute most in both designs, reflecting their role in label generation.

\subsection{Hidden Node Impact and Neural Network Redesign}
\label{subsec:exmodifiedxor}

\begin{table}[!th]
 \centering
 \caption{ $M$ vs. $\Pi(M)$ in five designs} 
 \label{tab:exmodifiedxor}
\begin{tabular}{llllllr}
   \\ \hline
   Design\: & \makecell{Hidden} \: & Epoch \: & \makecell{Batch} \: &
  \multicolumn{3}{c}{Accuracy ($\%$)}  \\
        \cmidrule{5-7} 
    &nodes & size & & $M$ &\:\: & $\Pi(M)$  
     \\ \hline
     1  & 128 &  200 & 16 & 91.9 & & 83.4   \\
      2  & 16 & 200  & 16 & 90.8 & & 82.6   \\
   3  &   64 & 200  & 16 & 92.3 & & 82.4   \\
    4  &  32 &  200 & 32 & 90.3 & & 80.7   \\
  \textbf{5}    & \textbf{10}  & 200 & 16 & \textbf{92.1} & & \textbf{82.8}   
   \\ \hline
    \end{tabular}
\end{table}

Similar to the first experiment, we create a baseline for our experiment using Weka. 
Since the dataset is more complex, the baseline has the accuracy of only 52.3\%. 
We report the top four designs for \dnn{} that achieves the accuracy better than 90\% (Lines 1-4, Table~\ref{tab:exmodifiedxor}).
The  rule extraction accuracy of \xdnnasp{} ranged from 80.7$\%$ (32 nodes) to 83.4$\%$ (128 nodes). 
A special design (\#5) uses only 10 nodes but boasts the $2^{nd}$ highest accuracy in both  trained network ($M$) and logic program $\Pi(M)$. This choice is due to the  
discovery of hidden node impact, discussed next.

\begin{figure}[!ht]
    \centering
    \includegraphics[width=.95\linewidth]{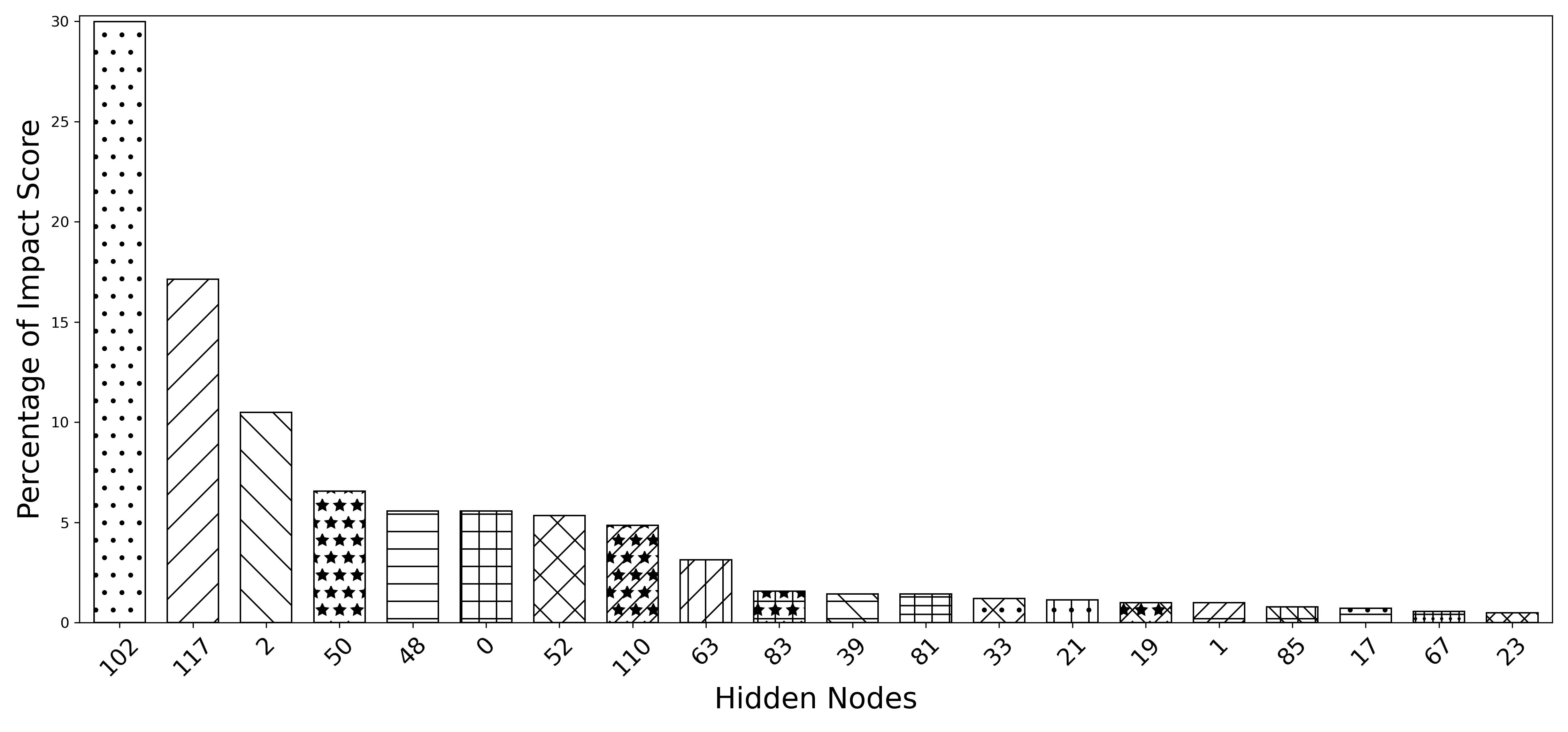}
    \caption{Top 20 hidden nodes' impact in fold 1, design 1 (Table \ref{tab:exmodifiedxor})}
    \label{fig:modifiedxorhiddenunits}
\end{figure}

Similar to feature importance, we define the impact score of a hidden node $j$ at layer $i$, denoted by $impact(h_i^j)$, as the product of two numbers based on its role: (a) the number of occurrences of atoms 
$h/5$ with $i$ and $j$
in the head which  indicates how much it is influenced by others, and (b) the number of occurrences of such atoms in the body which indicates its influences on other atoms.
Figure \ref{fig:modifiedxorhiddenunits} ranks the top 20 hidden nodes by impact score percentage, calculated by 
($ {impact(h_i^j)}/{\sum_{j=1}^{n} impact(h_i^j)}*100$), where $n=128$ 
for Fold 1 of Design 1 (Table \ref{tab:exmodifiedxor}).
It shows that only a small subset of nodes in the hidden layers significantly contributes to the output of the network. 
This trend is also observed in all other configurations.
This stipulates us to reduce the number of hidden nodes in Design 1 from 128 to 10 for an experiment. As we can see, this does not result in a reduction of accuracy of the network or the output of \xdnnasp{} (Li. 5, Tab. \ref{tab:exmodifiedxor}).  

\section{Conclusion}
\label{sec:conclusion}
 
We propose a system for understanding \dnn{}, \xdnnasp{}, 
that extracts answer set programs from trained models.
The extracted program maintains a high-level of accuracy of the original model and 
allows for the identification of features that are important to the prediction of the network 
that is consistent with statistical method.  
In addition, the program provides means for the evaluation of impacts of hidden nodes on the output of the network, which can be used to simplify the design of the original network. Our experiments show that  \xdnnasp{} performs better than previously developed decompositional systems for understanding \dnn{} such as 
\eclaire{}, \deepred{}, and \remD{} in several aspects. We note that the size of the program generated by \xdnnasp{} is linear in the size of the network. It might therefore be challenging if one wants to visually inspect the program. However, this is reasonable for analyzing purpose such as identifying important features or studying impacts of a hidden node or the interactions between the nodes or using the program to generate explanations for the outputs as in \cite{alviano2024xai}.
Future research could focus on reducing computational costs.
One potential approach is to examine the impact of groups of hidden nodes rather than individual ones. 
Another interesting direction is to explore specialized neural network architectures, such as Convolutional Neural Networks, to tackle  classification problems involving non-numerical data, such as images.

\section*{Acknowledgments}
Ly Trieu is supported by Los Alamos National Laboratory under the award number 047229. Son Tran was also partially supported by NSF grants 1914635 and 2151254.  

\bibliographystyle{eptcs}
\bibliography{main,bibtex}

\end{document}